\documentclass{article}

\usepackage{nips12submit_e,times, algorithm,algorithmic ,amsmath,amssymb,amsthm,amsfonts,subfigure,graphicx,hyperref,natbib}

\usepackage{cse599s12sp}

\title{A new boosting algorithm based on dual averaging scheme}

\author{
Nan Wang \\
LinkedIn Corporation \\
Mountain View, CA 94043 \\
\texttt{nawang@linkedin.com} \\
}

\nipsfinalcopy

\begin{document} 
\maketitle

\begin{abstract}
The fields of machine learning and mathematical optimization increasingly intertwined.  The special topic on supervised learning and convex optimization examines this interplay. The training part of most supervised learning algorithms can usually be reduced to  an optimization problem that minimizes a loss between model predictions and training data. While most optimization techniques focus on accuracy and speed of convergence, the qualities of good  optimization algorithm from the machine learning perspective can be quite different since machine learning is more than fitting the data. Better optimization algorithms that minimize the training loss can possibly give very poor generalization performance. In this paper, we examine a particular kind of machine learning algorithm, boosting,  whose training process can be viewed as functional coordinate descent on the exponential loss. We study the relation between optimization techniques and machine learning by implementing a new boosting algorithm. DABoost, based on dual-averaging scheme and study its generalization performance. We show that DABoost, although slower in reducing the training error, in general enjoys a better generalization error than AdaBoost.
\end{abstract}

\section{Introduction}
Optimization formulations and methods lie at the heart of many machine learning algorithms~\cite{murphy2012machine}.  A larger number of machine learning algorithms reduce to optimization problems. For example, support vector machine (SVM)~\cite{hearst1998support, scholkopf1998support, steinwart2008support} minimizes the hinge loss function between the training data and the model prediction. Latent models for sequential data~\cite{rabiner1986introduction, bahl1986maximum, huang2011predictive, Huang14} maximizes the conditional likelihood of observed data.  Logistic regression minimized the negative log conditional likelihood of training data given the model. Decision making problems~\cite{boutilier2002pomdp, levin1998using, Huang12, Huang13} can be formulated in terms of maximizing the sum of future rewards. Optimization algorithms are widely used for training a machine learning model. However, machine learning is more than simply a consumer of optimization techniques since machine learning concerns not only about model training but also model validation. The criterion used to validate the efficacy of a model is not the same as the criterion used for training the model. In a optimization problem, the quality of a good solution would be measured by its speed (convergence rate) and accuracy (objective gap). But in machine learning,  the generalization performance is perhaps the most important metric of solution quality. An optimization algorithm that has a poor convergence rate may score a  high generalization performance when it's applied to train a machine learning problem. Therefore machine learning presents new challenges to mathematical optimization. It is still an open question on what are the desirable properties of an optimization algorithm from the machine learning perspective. In this paper, we study boosting~\cite{freund1997decision, freund1999short}, a machine learning method that is famous for its resistance to over-fitting. For example, the winners of the HiggsML Challenge on Kaggle, develop and use the Boosting library, XGBoost~\cite{Chen:ICML2013}, to win this competition. We formulate boosting as an optimization problem on the exponential loss and show its equivalency to gradient descent. We introduce a novel variant of boosting algorithm, based on a new optimization algorithm called the dual averaging methods that minimizes the same exponential loss function. We examine the performance of the standard boosting algorithm and ours, from both an optimization perspective and a machine learning perspective. 

Boosting is a general method to derive strong learner from weak learning algorithms. The boosting method is based on the observation that finding  base (weak) learners that performs just slightly better than random guessing can be a lot easier than finding a single, highly accurate one. \cite{Kearns88} first postulated the conjecture of whether a combination of base learners can be boosted into an arbitrary accurate strong learner in the framework of PAC(probably approximately correct) learning model. \cite{Freund97} introduced the first practical boosting algorithm in binary classification,  called AdaBoost,  which repeatedly calls a base learning algorithm to train different classifiers that fits the re-sampled training examples from a different distribution. At each round the AdaBoost algorithm assigns larger weights on the harder examples, this effectively forces the base learning algorithm to focus its attention on the examples that were misclassified by the preceding classifier, and to come up with a new  classifier that is hopefully more accurate. AdaBoost then combines those weak classifier  by simply taking a weighted majority vote of their predictions. AdaBoost is shown to  give significant accuracy improvements over base learning algorithms. 

Looking to extend upon the success of AdaBoost, many attempts have been successfully at providing general algorithms for boosting. ~\cite{Breiman99} and \cite{Mason99} made the crucial links between AdaBoost and optimization by reformulating AdaBoost from a gradient descent point of view on an exponential loss function. This intuitive connection was further developed and many variants of AdaBoost were seen as performing gradient descent, but with different loss functions and different gradient descent methods. Furthermore, as pointed out by \cite{Grubb11} in their recent work,  the existing gradient-based boosting algorithm can fail to converge on some non-smooth convex objectives. To address this issue, they presented new algorithms that can be extended to arbitrary convex loss functions with convergence guarantee.  However, one limitation of these existing algorithms is that they computed new classifier based only on the sub-gradient of loss function at previous iteration. It was known that~\cite{Xiao10} this kind gradient descent method lacks the capability in exploiting the feasible set, especially when the loss function has additional regularization term such as $l_1$ norm for promoting sparsity. In this project, we would like to apply gradient descent method that involves the running average of all past sub-gradients of loss function (known as dual averaging method~\cite{Nes09,Baes11}), to the boosting framework. In addition, we would like to study the convergence results of the proposed algorithm. Finally, we will demonstrate experimental results that support our analysis and examples that show the need for the new algorithm based on dual-averaging scheme.

The remainder of this paper is organized in the following way. Section~\ref{sec:adaboost} first describe the Adaboost algorithm and formulated  it as a gradient descent on the exponential loss.  In section~\ref{sec:DA} we introduce the dual averaging method and show how to implement it in the boosting setting. We compare the performance results of both boosting algorithms in section~\ref{sec:results}. We conclude with discussions in section~\ref{sec:discussion}.

\section{AdaBoost}
\label{sec:adaboost}

\subsection{Algorithm description}

\begin{algorithm}
  \caption{AdaBoost}\label{algm:adaboost}
  \begin{algorithmic}
    \STATE Initial $D_1(i) = \1/n$  $\forall i = 1, \ldots, n$.
    \FOR{$t = 1,\ldots, T$}
    \STATE Train the weak classifier $h_t$  with smallest training error $\epsilon_t$ with respect to $D_t$.
 \STATE Choose $\eta_t = \frac{1}{2}\log\frac{1-\epsilon_t}{\epsilon_t}$
\STATE Let $Z_t = 2\sqrt{\epsilon_t(1 - \epsilon_t)}$ be a normalization factor so that $D_{t+1}$ will be a distribution, 
\STATE  Update $D_{t+1}(i) = D_t(i)\exp(-\eta_ty_ih_t(x_i))/Z_t$, for all $i$ 
\STATE The final classifier $f_t = f_{t-1} + \eta_t h_t = \sum_{s=1}^t \eta_sh_s$
    \ENDFOR
  \end{algorithmic}
\end{algorithm}

The AdaBoost algorithm (shown in Algorithm~\ref{algm:adaboost}) is arguably one of the most crucial  developments in  machine learning in the past two decades. AdaBoost can train  classifiers with extreme small generalization errors from base learners as weak as decision stumps or as strong as neural networks. Let $\{(x_i, y_i)\}_{i=1,\ldots, n}$ be the training set where the training instance $x_i \in \Xscr$ and the training label $y_i \in \{-1, 1\}$.   AdaBoost calls a given weak or base learning algorithm repeatedly in a series of rounds $t = 1,\ldots ,T$. Adaboost maintain a distribution $D_t$ over the training set, where $D_t(i)$ represents the weight of this distribution on training example $i$ on round $t$. Initially the distribution is uniform and all weights are set equally. But on each round, AdaBoost increase the weights of incorrectly classified examples by the previous classifier and decrease the weights of correctly classified ones. In this way, the weak learning algorithm is forced to focus on the hard examples in the training set. The job of the weak learner is to find a classifier $h_t$ that minimized the training error $\epsilon_t$ with respect to distribution $D_t$: 
\begin{align}
\epsilon_t = {\rm E}_{i \sim D_t}[y_t \neq h_t(x_i)] =  \sum_{i: h(x_i)\neq y_i} D_t(i)
\end{align}
In practice, the weak learner may be an algorithm that can make use of the weights $D_t$ on the training examples, or a subset of the training examples that are re-sampled according to $D_t$.  Once the weak classifier has been trained, AdaBoost chooses a parameter $\eta_t \frac{1}{2}\log\frac{1-\epsilon_t}{\epsilon_t}$ measures the importance that is assigned to $h_t$. Note that $\eta_t \ge 0$ if $\epsilon_t \le 1/2$, and the smaller $\epsilon_t$ gets the larger $\eta_t$ becomes. The final classifier $f_T$ is a weighted majority vote of $T$ weak classifiers where $\eta_t$ is the weight assigned to $h_t$.

\subsection{A gradient descent view}
Here we describe the general boosting algorithm as gradient descent in function space.  We consider the function $f: \Xscr \to \{-1,1\}$ in the function space $L^2(\Xscr, \mu)$ whose Lebesgue integral $\int_{\Xscr} \|f(x)\|^2 d\mu $ is finite.  The domain $\Xscr$ is measurable and $\mu$ is a probability measure $\hat{P}$ with empirical probability distribution estimated from training instances $\{x_i\}_{i = 1,\ldots n}$. The inner product  in this Hilbert space can be written as:
\begin{align}
  \label{eq:innerProdcut}
  \langle f, g \rangle_{\hat{P}} = \frac{1}{n}\sum_{i=1}^n \langle f(x_i), g(x_i)\rangle.
\end{align}
This definition of $f$ represents a great variety of machine learning algorithms ranged from multi-layer perceptron to decision tree.
Under the framework of empirical risk minimization, we would like to employ the gradient descent algorithm to minimize the empirical risk of $f$, which is a functional $\Rscr_{\rm emp} : L^2 \to \Re$:
\begin{align}
  \label{eq:empR}
  \Rscr_{\rm emp}[f] = \frac{1}{n} \sum_{i = 1}^n l(f(x_n),y_n).
\end{align}
where $l$ is the loss function that measures the difference between the prediction $f(x_n)$ and true label $y_n$. The gradient of $\Rscr_{\rm emp}$ with respect to a function $f$ is another function $g$ that makes $\Rscr_{\rm emp}[f + \eta g]$ change the most rapidly:
\begin{align}
  \label{eq:nablaR}
 g(x) =  \nabla \Rscr_{\rm emp}[f](x) = \frac{\partial \Rscr_{\rm emp}[F + \eta 1_x]}{\partial \eta} \mid \eta = 0
\end{align}
where $1_x$ is the indicator function of $x$. In contrast to the standard gradient descent algorithm, boosting restrict a set of allowable descent directions called the feasible set $\Hscr$, which correspond directly to the set of hypotheses generated by the base learner. $\Hscr$ can be a set of all possible decision trees generated by C4.5 algorithm, or a set of all possible support vector machines.  Given $\Hscr$, we would like to find a hypothesis $h^\star$ that is the closest to the computed negative gradient. $h^\star$  can then be found by projecting the negative gradient onto $\Hscr$. 
\begin{align}
  \label{eq:hstar}
  h^\star = \arg\max_{h\in \Hscr} \langle - \nabla \Rscr_{\rm emp}[f], h \rangle_{\hat{P}}.
\end{align}
Finally the gradient descent algorithm will chose a step size $\eta_t$ such that the empirical risk at the updated function $\Rscr_{\rm emp}[f + \eta_t h^\star]$  is minimized.
\begin{algorithm}
  \caption{Gradient Projection Algorithm}
\label{algm:gd}
  \begin{algorithmic}
    \STATE Given starting point $f_0$.
    \FOR{$t = 1, \ldots, T$.}
    \STATE Compute the gradient $ \nabla \Rscr_{\rm emp}[f_{t-1}]$
    \STATE Find $h^* = \arg\max_{h\in \Hscr} \langle - \nabla \Rscr_{\rm emp}[f_{t-1}], h \rangle_{\hat{P}}. $
   \STATE Find a step size $\eta_t = \arg\min_{\eta} \Rscr_{\rm emp}[f_{t-1} + \eta h^\star]$
   \STATE Update $f_{t} = f_{t-1} + \eta_t h^\star$
    \ENDFOR
  \end{algorithmic}
\end{algorithm}

For Adaboost, it can be shown that  $\Rscr_{\rm emp}[f] = \sum_i \exp(-f(x_i)y_i)$. We have 
\begin{align}
  \nabla \Rscr_{\rm emp}[f](x) = - y_i \exp(-y_if(x_i)) \delta_{x, x_i}
\end{align}
where $\delta_{x,x_i} = 1$ if $x = x_i$, otherwise $\delta_{x,x_i} = 0$. Finding the closet hypothesis $h^\star$ from $\Hscr$ is equivalent to maximizing
\begin{align*}
 \langle - \nabla \Rscr_{\rm emp}[f_{t-1}], h \rangle_{\hat{P}} &= - \frac{1}{n} \sum_i -y_i h(x_i) {\rm e}^{(-y_i f_{t-1}(x_i))} \\
&=  \sum_i y_i h(x_i) D_t(i) [\prod_{s=1}^{t-1}Z_s] \\ &\propto [1 - 2\epsilon_t ]
\end{align*}
 where  $D_1(i) = \frac{1}{n}$,  $\epsilon_t = {\rm E}_{i \sim D_t}[y_t \neq h_t(x_i)]$ and
\begin{align}
 D_t(i) &= \frac{\exp(-y_i f_{t-1}(x_i))}{n\prod_{s=1}^{t-1}Z_s} \nonumber \\
&= \frac{D_{t-1}(i)}{Z_{t-1}} \exp(- \eta_{t-1} y_i h^\star(x_i)) \\
Z_t &= 2\sqrt{\epsilon_t(1-\epsilon_t)}
\end{align}
This projection step is equivalent to finding a base hypothesis with smallest  mis-classification error $\epsilon_t$ over a boosted training set with distribution $D_t$. Finally, we choose the step size $\eta_t$ such that
\begin{align}\label{eq:updateEta}
  \frac{d \Rscr_{\rm emp}[f_{t-1} + \eta h]}{d\eta} & = 
 -\sum_{i=1}^n y_i h(x_i) \exp(-y_i f_{t-1}(x_i) -y_i h(x_i)\eta) \\
&\propto -\sum_{i=1}^n y_i h(x_i) D_t(i) \exp(-y_i h(x_i)\eta) \\
& = \epsilon_t \exp(\eta) - (1 - \epsilon_t) \exp(\eta) = 0\\
\eta_t &= \frac{1}{2}\log\frac{1-\epsilon_t}{\epsilon_t}
\end{align}
Therefore, with exponential loss function, the gradient projection algorithm~\ref{algm:gd} is equivalent to the AdaBoost algorithm ~\ref{algm:adaboost}. By viewing AdaBoost as a gradient descent in the functional space, it's tempting to conclude that AdaBoost a just an algorithm for minimizing exponential loss and more (less) powerful optimization techniques for the same loss should work even better(worse).  In the next sections, we are going to test this conclusion by introducing a new variant of boosting algorithms that implements a different optimization technique.

\section{Boosting with  dual averaging method}\label{sec:DA}

  \begin{algorithm}
  \caption{Dual Averaging Algorithm}\label{algm:da}
  \begin{algorithmic}
  \STATE Given objective function $R[f]$  and regularization function $d(f)$.
\FOR{$t= 1, \ldots, T$}
  \STATE Compute gradient $g_t = \nabla_f R[f_t]$
  \STATE Choose $\lambda_t > 0$, set $s_{t+1} = s_t + \lambda_t g_t$
  \STATE Choose $\beta_{t+1}$. Set $$h^\star = \pi_{\beta_{t+1}}( -s_{t+1})$$
    where $$\pi_\beta(s) = \arg\min_{h\in\mathcal{H}}\{ \langle s,h\rangle + \beta d(h) \} $$
\ENDFOR
\STATE Update $f_{t+1} = f_t + \alpha_{t+1} h^\star$
  \end{algorithmic}
  \end{algorithm}
Dual averaging method (shown in algorithm~\ref{algm:da}) has recently been introduced in convex optimization by \cite{Nes09}.  In the paper of \cite{Baes11}, they proposed an alternative viewpoint of the Hedge algorithm using dual averaging method. The hedge algorithm has been known for its close relation to the AdaBoost algorithm. However, the hedge algorithm and the AdaBoost  algorithm differ in many different ways. First, in hedge algorithm (see \cite{Baes11} for more details), the weight  $D_t(i)$ increases if $i$th strategy is a ``good'' action at round $t$, while in AdaBoost the weight  $D_t(i)$ increases if the $t$ hypothesis suggests a ``bad'' prediction on the $i$th example. The loss in hedge algorithm that measures the success  of the strategy is a actually a measurement of hardness of an example in AdaBoost.  Thus, the algorithm that minimizes the loss in Hedge algorithm will not minimized the training error $\epsilon_t$ in AdaBoost. Secondly yet more importantly, in AdaBoost the updating rule for the weights is $D_{t+1}(i) \propto D_t(i) \exp(-\eta_t)^{y_i h(x_i)}$ with a time-varying parameter $\exp(-\eta_t)$ that changes at each iteration according to training error $\epsilon_t$. But in hedge algorithm, this parameter is fixed ahead of time.\footnote{the parameter is denoted by $\gamma$ in ~\cite{Baes11}} Therefore, the algorithm described in~\cite{Baes11} can not directly applied to a boosting algorithm. Instead, we need to design a novel dual averaging algorithm that is based on boosting setting.

To correctly apply the dual averaging method in the boosting setting, we need to define the dual variable $s_t= \sum_{k=1}^t \lambda_k g_k$ in the functional space:
\begin{align}
  s_t =  \sum_{k=1}^t - \lambda_k  y_i \exp(-y_if(x_i)) \delta_{x, x_i}.
\end{align}
Because the hypothesis class $\Hscr$ applies to arbitrary weak learner. It's not clear how to define a regularization function of rule-based learner (like zero-R) and tree-based learner (such as decision stumps and CART). We let $R(h) = 0$, then finding a weak classifier $h^\star$ can be written as:
\begin{align}
      h^\star &=  \arg\min_{h\in\mathcal{H}} \sum_i \frac{ y_i h(x_i)}{m}  \sum_{k=1}^{t-1} \lambda_k \exp[-y_i f_k(x_i)] \nonumber \\
 &=  \arg\min_{h\in\mathcal{H}} \sum_i \frac{ y_i h(x_i)}{m}  D_{t}(i)\nonumber \\
  &= \arg\min_h E_{i\sim D_{t}} [1 - 2  \epsilon_{t}] 
\end{align}
where $\epsilon_t = {\rm E}_{i \sim D_t}[y_t \neq h_t(x_i)]$ the probability (weight) for each instance
\begin{align}\label{eq:dtinDA}
    D_{t}(i) = \frac{ \sum_{k=1}^{t-1} \lambda_k \exp[-y_i f_k(x_i)]}{\sum_{i=1}^n \sum_{k=1}^{t-1} \lambda_k \exp[-y_i f_k(x_i)] }
\end{align}

  \begin{algorithm}
  \caption{DABoost}\label{algm:daboost}
  \begin{algorithmic}
  \STATE  Given training samples $(x_1, y_1), \ldots, (x_n, y_n)$
  \STATE  Initialize $D_1 = \frac{1}{n}$
  \FOR{$t = 1, \ldots, T$}
    \STATE Re-sample training data  from $D_t$.
    \STATE Find the  weak classifier $h_t$ that minimize the error rate $\epsilon_t$.
\item Set the step size   $\alpha_t = \frac{1}{2}\sqrt{\frac{1-\epsilon_t}{\epsilon_t}}$
      \STATE Update distribution $D_{t+1} \propto  \sum_k \lambda_k \exp[-y_i f_k(x_i)]$
\ENDFOR
\STATE Final classifier $f(x) = \sum_{t} \alpha_t h_t(x)$
  \end{algorithmic}
  \end{algorithm}
Equation~\ref{eq:dtinDA}  defines a way to update the distribution $D_t$ in dual averaging setting. And we choose the time step $\eta_t$ as in equation~\ref{eq:updateEta}, $\eta_t = \frac{1}{2}\log(\frac{1-\epsilon_t}{\epsilon_t})$. Finally we introduce a novel boosting algorithm, called DABoost (shown in algorithm~\ref{algm:daboost}), based on dual averaging method.

\begin{figure}
  \centering
  \subfigure[]{
\includegraphics[width=0.4\textwidth]{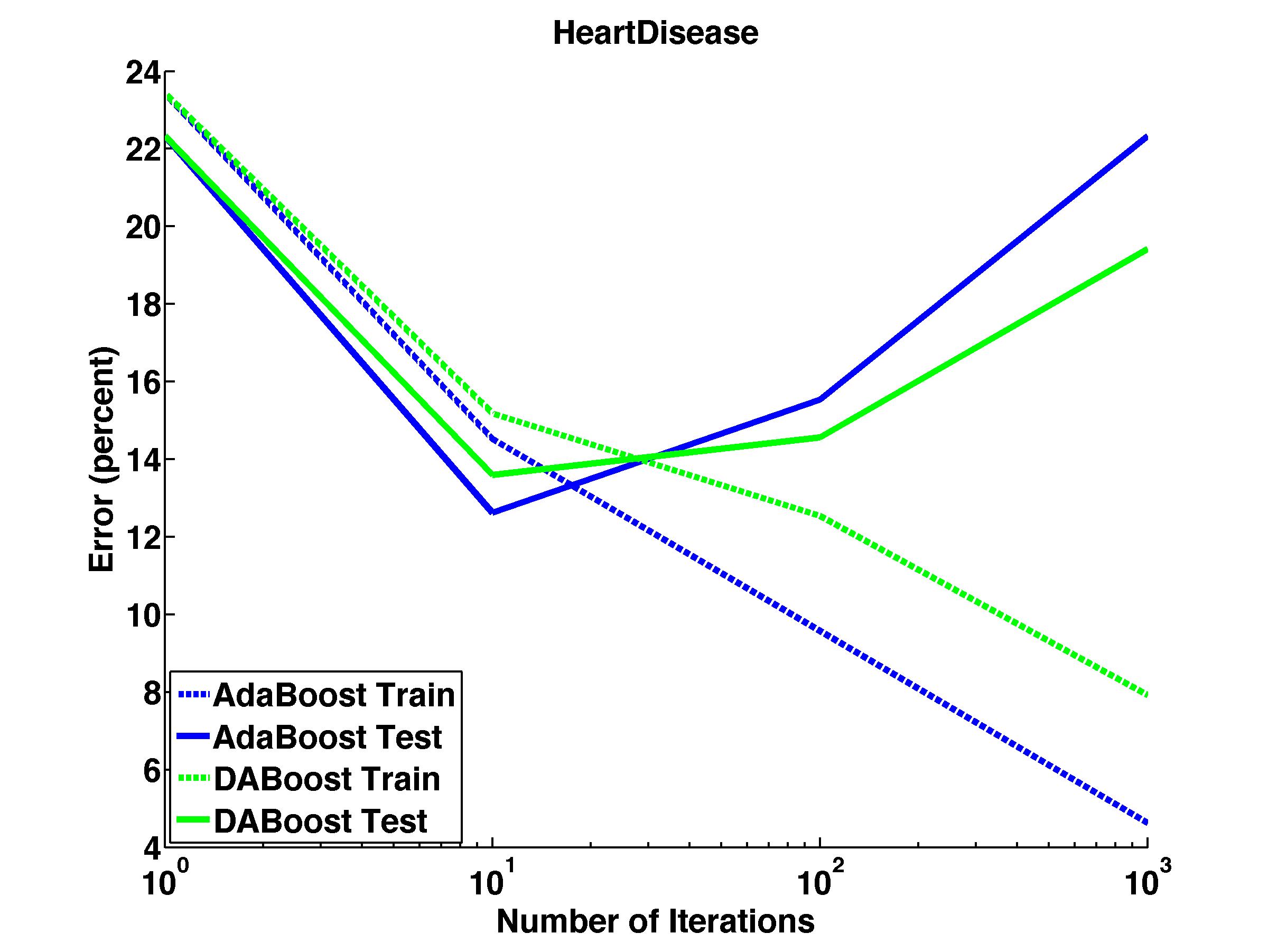}
\label{fig:heart}
  }
 \subfigure[]{
\includegraphics[width=0.4\textwidth]{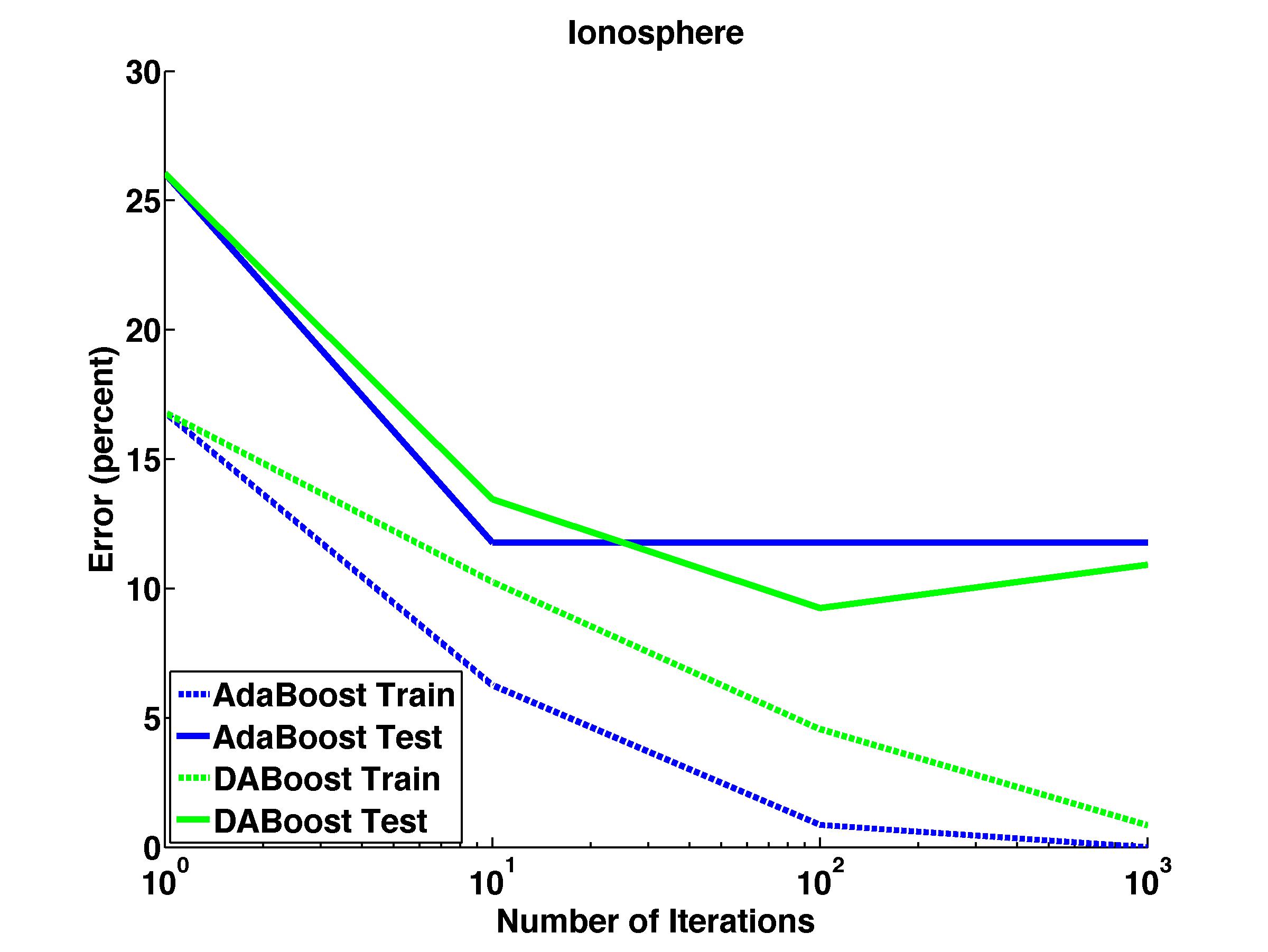}
\label{fig:ion}
  }
  \label{fig:results}
\caption{This figure shows the Training and test errors  of AdaBoost and DABoost, both using stumps as base learner, on data sets (a)Heart Disease, (b)Inonospear  DABoost tends to produce classifiers wit higher bias but less variance.}
\end{figure}

Based on algorithm~\ref{algm:daboost}, we implement the DABoost algorithm in the WEKA environment so that our DABoost algorithm can call basically any existing machine learning algorithms as the base learner. We fix the time-dependent importance parameter $\lambda_t = 1$ be constant in the implementation.

\section{Results}\label{sec:results}

In this section, we evaluate our DABoost algorithm using different data sets in UCI machine learning repository~\cite{Lichman2013}, and compare our training error and test error to those of AdaBoost. Because the training port of both AdaBoost and DABoost can be viewed as convex optimization on the exponential loss. The training error represents the objective gap of the loss function, a measurement of quality  from the optimization perspective, while the test error represent the generalization performance that is a measurement of quality from  the machine learning perspective. 

We start with a toy example where the instance $x$ is drawn uniformly from $[-1, 1]^{100}$, and the label $y$ is the majority vote of three coordinates. The size of the training set $n = 1,000$. We use this data set to test the correctness of our DABoost algorithm. With this simple data set, both DABoost and AdaBoost (boosting stumps) achieve  $0\%$ training error after three iterations, as expected. In comparison, popular machine learning algorithms such as SVM, logistic regression and multilayer perceptron only score test errors greater than $15\%$ (note that tree-based algorithms such as CART and C4.5 can also score $0\%$ test error).

Figure~\ref{fig:heart} shows the performance of AdaBoost and DABoost, both using stumps as base learner, in Heart Disease data set. The blue curves represent the results from Adaboost while the green curves represent the results from DABoost. Training errors are shown in dashed lines while test errors are shown in solid lines. The heart disease data set contains $14$ attributes. The label refers to the presents of heart disease in the patient. The heart disease data set is one of few training sets on which the AdaBoost algorithm overfits. As shown in the figure, DABoost converges slowly in terms of training error but suffers less over-fitting in terms of test error.

Figure~\ref{fig:ion} shows the performance of AdaBoost and DABoost, both boosting decision stumps, in ionosphere data set  The ionosphere data set contains $34$ continuous attributes. The label is either ``good'' or ``bad'' where ``good'' radar returns are those showing evidence of some type of structure in the ionosphere and  ``bad'' returns are those that do not. The ionosphere  data set is commonly used in the machine learning literature. As shown in the figure, DABoost again converges slower than Adaboost in terms of training error but achieves better test error.

In general, DABoost gets higher training error but enjoys lower test error. Similar behavior have been observed in many other data sets such as the letter data set by boosting a C4.5 base learner, and in the diabetes data set by boosting stumps.  This is due to the bias-variance trade-off. DABoost update the distribution $D_t$ based on the an average of loss over all previous iterations. The resulting classifier becomes less flexible because as the number of iterations $t$ increases, the distribution $D_t$ changes slowly due to the update rule in equation~\ref{eq:dtinDA}. Thus, DABoost tends to produce classifiers wit higher bias but less variance.

\begin{figure}
  \centering
\includegraphics[width=0.4\textwidth]{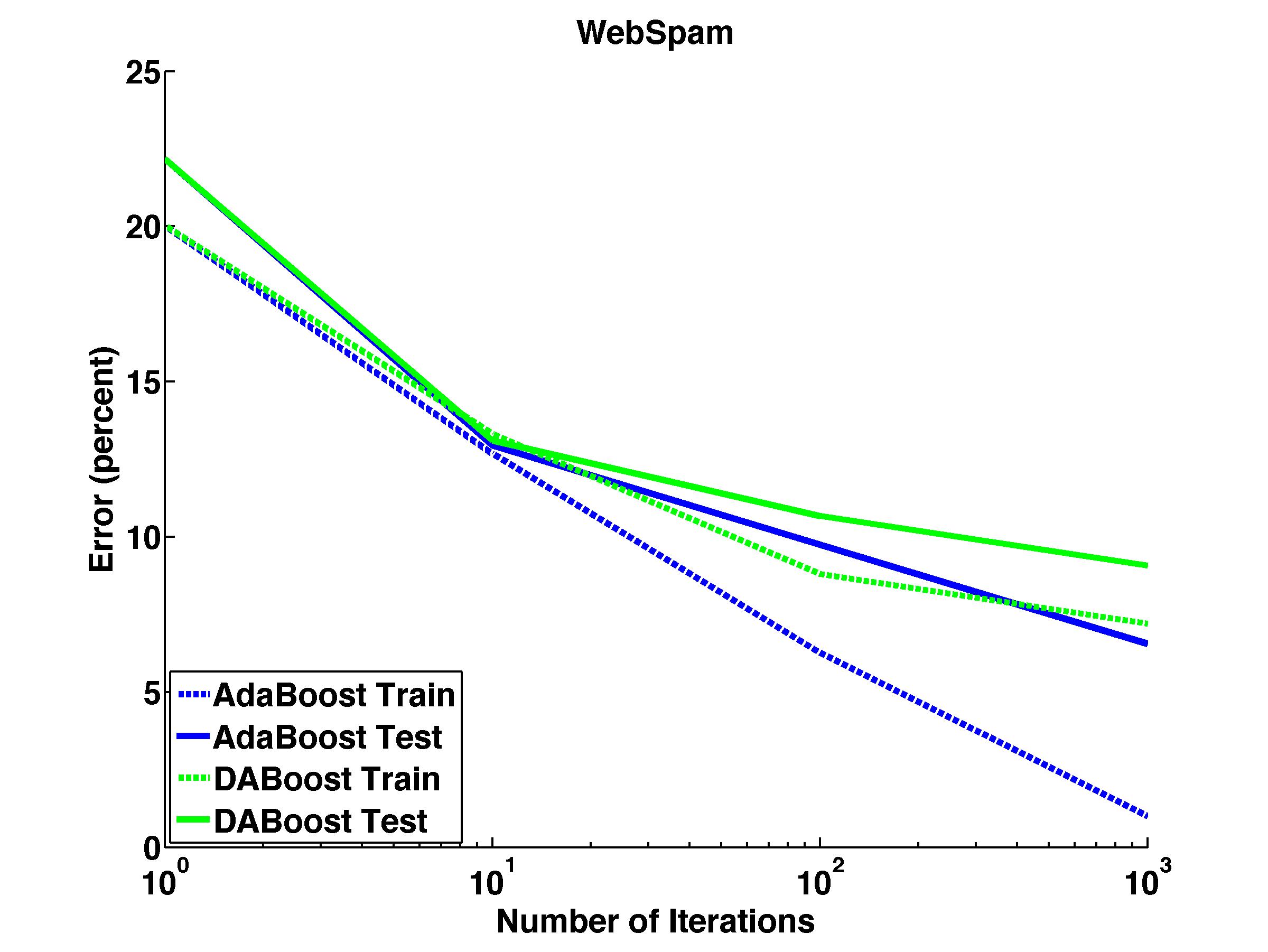}
\label{fig:spam}
\caption{This figure shows the Training and test errors  of AdaBoost and DABoost, both using stumps as base learner, on data sets (a)Heart Disease, (b)Inonospear  DABoost tends to produce classifiers wit higher bias but less variance.}
\end{figure}

However, DABoost doesn't always have better generalization performance than Adaboost.  Figure~\ref{fig:spam} shows the performance of AdaBoost and DABoost, whose base learner is decision stumps, in Webb Spam Corpus data set.  Web spam is defined as Web pages that are created to manipulate search engines and deceive Web users.  All positive examples were taken and the negative examples were created by randomly traversing the Internet starting at well known (e.g. news) web-sites. In this data set, any continuous one byte is treated as a word and the world count is used as the feature value. Each instance is normalize to unit length.  The number of total features is $254$.  Due to memory constraints, only $1\%$ of instances ($3,500$ instances) are used for training.   As shown in the figure, DABoost converges slower than AdaBoost in terms of training error but also suffers larger test error.  The reason for such poorer performance is due to the noise in the highly biased labels in the data set.

\section{Conclusion}\label{sec:discussion}
In this paper, we discuss the quality of a good optimization algorithm from both a machine learning perspective and a mathematical programming perspective. We postulated that a slower convergent optimization algorithm might result in a better machine learning algorithm with better generalization performance. We test this postulation by introducing a new variant of boosting algorithm, DABoost, based on dual averaging gradient descent method on exponential loss. Our simulation results show although slower in obtaining small training error, DABoost in general enjoys better generalization error than AdaBoost.  

Our implementation of DABoost is still far from complete and demands a series of future research.  We fix the time-dependent importance parameter $\lambda_t = 1$ be constant in the implementation.  A time-varying $\lambda_t$ might lead to different results. Moreover, we simplifies the dual averaging algorithm by restraining  the regularization function $d(h) = 0$. $l_1$ or $l_2$ regularization might be applied to base learners that can be parametrized by a vector of real numbers. 

DABoost is based on dual averaging algorithm, the recently introduced convex optimization algorithm that has similar linear convergence rate as the gradient descent. In our further work, more powerful optimization techniques such as accelerated gradient descent with quadratic convergence rate might be implemented in the boosting framework.

\bibliographystyle{icml2012}
 \bibliography{boosting}

\begin{thebibliography}{24}
\providecommand{\natexlab}[1]{#1}
\providecommand{\url}[1]{\texttt{#1}}
\expandafter\ifx\csname urlstyle\endcsname\relax
  \providecommand{\doi}[1]{doi: #1}\else
  \providecommand{\doi}{doi: \begingroup \urlstyle{rm}\Url}\fi

\bibitem[Baes \& Burgisser(2011)Baes and Burgisser]{Baes11}
Baes, M. and Burgisser, M.
\newblock Hedge algorithm and dual averaging schemes.
\newblock \emph{arXiv}, 1112\penalty0 (1275), 2011.

\bibitem[Bahl et~al.(1986)Bahl, Brown, De~Souza, and Mercer]{bahl1986maximum}
Bahl, LR, Brown, Peter~F, De~Souza, Peter~V, and Mercer, Robert~L.
\newblock Maximum mutual information estimation of hidden markov model
  parameters for speech recognition.
\newblock In \emph{proc. icassp}, volume~86, pp.\  49--52, 1986.

\bibitem[Boutilier(2002)]{boutilier2002pomdp}
Boutilier, Craig.
\newblock A pomdp formulation of preference elicitation problems.
\newblock In \emph{AAAI/IAAI}, pp.\  239--246, 2002.

\bibitem[Breiman(1999)]{Breiman99}
Breiman, L.
\newblock Prediciton games and acring algorithms.
\newblock \emph{Nueral Computation}, 11:\penalty0 1493--1517, 1999.

\bibitem[Chen et~al.(2013)Chen, Li, Yang, and Yu]{Chen:ICML2013}
Chen, Tianqi, Li, Hang, Yang, Qiang, and Yu, Yong.
\newblock General functional matrix factorization using gradient boosting.
\newblock In \emph{Proceeding of 30th International Conference on Machine
  Learning (ICML'13)}, volume~1, pp.\  436--444, 2013.

\bibitem[Freund \& Schapire(1997{\natexlab{a}})Freund and Schapire]{Freund97}
Freund, Y. and Schapire, R.~E.
\newblock A decision-theorectic generalization of on-line learning and an
  application to boosting.
\newblock \emph{Journal of Computer and System Sciences}, 55\penalty0
  (1):\penalty0 119--139, 1997{\natexlab{a}}.

\bibitem[Freund \& Schapire(1997{\natexlab{b}})Freund and
  Schapire]{freund1997decision}
Freund, Yoav and Schapire, Robert~E.
\newblock A decision-theoretic generalization of on-line learning and an
  application to boosting.
\newblock \emph{Journal of computer and system sciences}, 55\penalty0
  (1):\penalty0 119--139, 1997{\natexlab{b}}.

\bibitem[Freund et~al.(1999)Freund, Schapire, and Abe]{freund1999short}
Freund, Yoav, Schapire, Robert, and Abe, N.
\newblock A short introduction to boosting.
\newblock \emph{Journal-Japanese Society For Artificial Intelligence},
  14\penalty0 (771-780):\penalty0 1612, 1999.

\bibitem[Grubb \& Bagnell(2011)Grubb and Bagnell]{Grubb11}
Grubb, A. and Bagnell, J.~A.
\newblock Generalized boosting algorithms for convex optimization.
\newblock In \emph{Proceedings fo the 28th International Conference on Machine
  Learning}, Bellevue, WA, June 2011.

\bibitem[Hearst et~al.(1998)Hearst, Dumais, Osman, Platt, and
  Scholkopf]{hearst1998support}
Hearst, Marti~A., Dumais, Susan~T, Osman, Edgar, Platt, John, and Scholkopf,
  Bernhard.
\newblock Support vector machines.
\newblock \emph{Intelligent Systems and their Applications, IEEE}, 13\penalty0
  (4):\penalty0 18--28, 1998.

\bibitem[Huang \& Rao(2013)Huang and Rao]{Huang13}
Huang, Y. and Rao, R. P.~N.
\newblock Reward optimization in primate brain: A pomdp model of decision
  making under uncertainty.
\newblock \emph{PLoS One}, 8\penalty0 (1), 2013.

\bibitem[Huang et~al.(2012)Huang, Friesen, Hanks, Shadlen, and Rao]{Huang12}
Huang, Y., Friesen, A.~L., Hanks, T.~D., Shadlen, M.~N., and Rao, R. P.~N.
\newblock How prior probability influences decision making: A unifying
  probabilistic model.
\newblock \emph{Advances in Neural Information Processing Systems (NIPS)},
  2012.

\bibitem[Huang \& Rao(2014)Huang and Rao]{Huang14}
Huang, Yanping and Rao, Rajesh~P.
\newblock Neurons as monte carlo samplers: Bayesian inference and learning in
  spiking networks.
\newblock In Ghahramani, Z., Welling, M., Cortes, C., Lawrence, N.D., and
  Weinberger, K.Q. (eds.), \emph{Advances in Neural Information Processing
  Systems 27}, pp.\  1943--1951. Curran Associates, Inc., 2014.

\bibitem[Huang \& Rao(2011)Huang and Rao]{huang2011predictive}
Huang, Yanping and Rao, Rajesh~PN.
\newblock Predictive coding.
\newblock \emph{Wiley Interdisciplinary Reviews: Cognitive Science}, 2\penalty0
  (5):\penalty0 580--593, 2011.

\bibitem[Kearns \& Valiant(1988)Kearns and Valiant]{Kearns88}
Kearns, M.~J. and Valiant, L.~G.
\newblock Learning boolean formulae or finte automata is as hard as factoring.
\newblock Technical report, Department of Computer Science, Harvard University,
  1988.

\bibitem[Levin et~al.(1998)Levin, Pieraccini, and Eckert]{levin1998using}
Levin, Esther, Pieraccini, Roberto, and Eckert, Wieland.
\newblock Using markov decision process for learning dialogue strategies.
\newblock In \emph{Acoustics, Speech and Signal Processing, 1998. Proceedings
  of the 1998 IEEE International Conference on}, volume~1, pp.\  201--204.
  IEEE, 1998.

\bibitem[Lichman(2013)]{Lichman2013}
Lichman, M.
\newblock {UCI} machine learning repository, 2013.
\newblock URL \url{http://archive.ics.uci.edu/ml}.

\bibitem[Mason et~al.(1999)Mason, Baxter, Barlett, and Frean]{Mason99}
Mason, L., Baxter, J., Barlett, P., and Frean, M.
\newblock Functional gradient techniques for combining hypotheses.
\newblock In \emph{Advances in large margin classifiers}. MIT Press, Cambridge,
  1999.

\bibitem[Murphy(2012)]{murphy2012machine}
Murphy, Kevin~P.
\newblock \emph{Machine learning: a probabilistic perspective}.
\newblock MIT press, 2012.

\bibitem[Nesterov(2009)]{Nes09}
Nesterov, Y.
\newblock Primal-dual subgradient methods for convex problems.
\newblock \emph{Mathematical Programming}, 120\penalty0 (1):\penalty0 221--259,
  2009.

\bibitem[Rabiner \& Juang(1986)Rabiner and Juang]{rabiner1986introduction}
Rabiner, Lawrence~R and Juang, Biing-Hwang.
\newblock An introduction to hidden markov models.
\newblock \emph{ASSP Magazine, IEEE}, 3\penalty0 (1):\penalty0 4--16, 1986.

\bibitem[Sch{\"o}lkopf \& Smola(1998)Sch{\"o}lkopf and
  Smola]{scholkopf1998support}
Sch{\"o}lkopf, Bernhard and Smola, Alex.
\newblock Support vector machines.
\newblock \emph{Encyclopedia of Biostatistics}, 1998.

\bibitem[Steinwart \& Christmann(2008)Steinwart and
  Christmann]{steinwart2008support}
Steinwart, Ingo and Christmann, Andreas.
\newblock \emph{Support vector machines}.
\newblock Springer Science \& Business Media, 2008.

\bibitem[Xiao(2010)]{Xiao10}
Xiao, L.
\newblock Dual averaging methods for regularized stochastic learning and online
  optimization.
\newblock \emph{The Journal of machine learning research}, 11, 2010.

\end{thebibliography}

\end{document}